\documentclass{IOS-Book-Article}

\usepackage{mathptmx}
\usepackage{soul}\setuldepth{article}

\def\hb{\hbox to 11.5 cm{}}

\begin{document}

\pagestyle{headings}
\def\thepage{}

\begin{frontmatter} 

\title{The Unreasonable Effectiveness of the Baseline: Discussing SVMs in Legal Text Classification}


\author[A]{\fnms{Benjamin} \snm{Clavié}%
\thanks{Contact: Benjamin Clavié, Jus Mundi, 30 Rue de Lisbonne,
75008 Paris, France;
b.clavie@jusmundi.com.}}
and
\author[A]{\fnms{Marc} \snm{Alphonsus}}

\runningauthor{B. Clavié and M. Alphonsus}
\runningtitle{Discussing SVMs in Legal Text Classification}

\address[A]{Jus Mundi}

\begin{abstract}
We aim to highlight an interesting trend to contribute to the ongoing debate around advances within legal Natural Language Processing. Recently, the focus for most legal text classification tasks has shifted towards large pre-trained deep learning models such as BERT. In this paper, we show that a more traditional approach based on Support Vector Machine classifiers reaches surprisingly competitive performance with BERT-based models on the classification tasks in the LexGLUE benchmark. We also highlight that error reduction obtained by using specialised BERT-based models over baselines is noticeably smaller in the legal domain when compared to general language tasks. We present and discuss three hypotheses as potential explanations for these results to support future discussions.
\end{abstract}

\end{frontmatter}

\section{Introduction}
Recently, the state-of-the-art in many Natural Language Processing (NLP) tasks has been achieved by large pre-trained models such as BERT and its variants \cite{bert}. Specialised BERT-based models have been developed for many fields, establishing the state-of-the-art in domain specific tasks, as evidenced in the biomedical domain \cite{biobert}.

In legal NLP, recent work has focused on exploring the applications of BERT-based approaches on a variety of existing tasks and how to best adapt BERT to the legal domain \cite{legalbertacl, legalberticail}. These efforts, while successful at establishing state-of-the-art on a variety of tasks, also reveal an interesting trend: the performance gain between a general language BERT and a specifically legal-language trained BERT appears to be smaller than in other specialised domains \cite{legalberticail}.

A common application of legal NLP is text classification. Text classification tasks target various kinds of legal insight, such as predicting the outcome of a ruling from a decision's body \cite{echr}, whether a given clause is likely to be unfair to a customer \cite{claudette} or identifying the topic of a contract close \cite{ledgar}. Recently, the LexGLUE benchmark has been released to allow for easier, more transparent benchmarking in legal NLP \cite{lexglue}.

Little attention has been given to comparing these new BERT-based approaches to well-optimised baselines, such as Support Vector Machine (SVM)-based classifiers, which historically perform well on text classification tasks, opting instead for comparisons with other deep learning-based baselines.

In this short paper, we aim to (A) highlight the very strong performance of optimised \textit{baseline} classifiers on multiple legal text classification tasks compared to deep learning classifiers, (B) show that the gains from BERT-based approaches is noticeably smaller on legal-domain tasks than on general tasks and (C) discuss three hypotheses to explain the previous two phenomena.


\subsection{General Domain}

For all general domain tasks, we use results from BERT-ITPT-FiT \cite{bertclf}, which optimises BERT for text classification, on four common benchmarks. For SVM results, we report the score of the best performing variant from a large scale comparison \cite{baselines}.

\subsection{Legal Domain Experiments \& Baselines}

For ease of comparison and reproducibility, we benchmark SVM classifiers on LexGLUE \cite{lexglue}. LexGLUE, is a benchmark suite comprising of commonly studied legal NLP tasks. For the purpose of this paper, we evaluate our models on all six classification tasks and leave the QA task aside and focus on the six text classification tasks:

\textbf{ECtHR A and B} \cite{echr, rationale} are multi-label tasks which use \textit{facts} part of European Court of Human Rights (ECtHR) rulings and the goal is to predict which article(s) were found by the court to have been violated (A) and which article(s) were considered by the court as allegedly violated (B).

\textbf{SCOTUS} \cite{lexglue} is a multi-class task on opinions from the Supreme Court of the United States which need to be classified into one of 14 issue areas.

\textbf{EUR-LEX} \cite{eurlex} is a multi-label task on EU law documents. The labels are comprised of 100 common EuroVoc legal concepts. The aim of this task is to identify all relevant EuroVoc label linked with a document.

\textbf{LEDGAR} \cite{ledgar} is a multi-class task on contract provisions (paragraph) from the US Securities and Exchange Commission (SEC) where the goal is to detect the main topic of a given provision.

\textbf{Unfair Terms of Service (ToS)} \cite{claudette} is a multi-label task with clauses from the Terms of Services of 50 online plateforms. The aim of this task is to detect if a clause is likely to violate consumer rights and exactly which right is at risk of being violated. 

Further information about the LexGLUE tasks is provided by Chalkidis et al. \cite{lexglue}.\\
\\
On each of the legal tasks, we train and evaluate SVM classifiers with modest optimisation. Results are reported as the average of ten runs. We also experiment with NBSVM, an SVM classifier using Naïve Bayes features to represent words \cite{nbsvm}. We release the code used to train all models. \footnote{Code available at \href{https://gitlab.com/jusmundi-group/public/Legal-svm-baselines}{https://gitlab.com/jusmundi-group/public/Legal-svm-baselines}}. \\
\\
For BERT-based models, we report the benchmark results from the LexGLUE paper \cite{lexglue}. Following their nomenclature, we refer to the legal Legal-BERT model from Chalkidis et al. \cite{legalbertacl} as Legal-BERT and the Legal-BERT model from Zheng et al. \cite{legalberticail}, trained on US case law documents, as CaseLaw-BERT.

\subsection{Metrics and Evaluation}

In line with the nature of this paper, all metrics reported follow the existing literature. For all General Domain tasks, the metric used is accuracy over the test set. As these tasks can be multi-class but not multi-label classification tasks, accuracy is identical to micro-averaged F1 score. \cite{f1}

For all legal tasks, we report both micro-averaged (µF1) and macro-averaged (mF1) F1 score for all tasks to allow for easier comparison with both other models evaluated on LexGLUE and results from general domain tasks.

In all cases, we report the error reduction between the best performing BERT-based model and SVMs as the percentage decrease in error rate between models to simplify evaluating the impact of using a different model over multiple tasks. The error rate is calculated as \begin{math}{\textit{100}-Score}\end{math}.

\section{Classification Results}
Table~1 gives an overview of the various classification results on General Domain tasks and presents the error reduction obtained by using the a fine-tuned BERT model over the best SVM classifier in the literature.

Table~2 presents the classification results on the 6 LexGLUE classification tasks. We report the results for our best performing SVM as well as the results from BERT-Base, Legal-BERT and CaseLaw-BERT from the existing LexGLUE benchmark. We also compute the error reduction between the SVM classifier and the best performing BERT-based model for each task.

The error reduction between SVM and BERT models in the general domain is  high, at \textbf{85.1\%} on average over the four tasks, with the lowest reduction being 77.2\%.

The difference is much less stark within the legal domain. BERT-based models do obtain the best results on all six takss, with Legal BERT models reaching the best performance on five out of the six tasks and a general domain BERT slightly outperforming them on \textbf{ECtHR (A)}. However, in all cases, the performance increase is much less pronounced than in other domains, with an average micro-F1 error reduction across all six tasks of just \textbf{18.1\%}.

The average reduction in macro-F1 is overall similar, with an average reduction of \textbf{18.3\%}. However, is noticeable that this reduction in is considerably more pronounced on both \textit{ECtHR} tasks. The ECtHR dataset is rather imbalanced, with all both tasks containing labels with very few examples and others covering a large proportion of the data \cite{lexglue}. This seems to point towards the BERT models' ability t oget better performance with lower quantities of data.

\begin{table}[]
\caption{Results for the best performing model of each kind on a variety of General Domain (GD). \textit{Error reduction is calculated between the Best SVM and the best BERT variant.}}
\begin{tabular}{|c|c|c|c|c|c|}
\hline
\textbf{Model}           & \textbf{AGNews} & \textbf{IMDB} & \textbf{Yelp!} & \textbf{DBPedia} & \textbf{Average} \\ \hline
\textbf{Best SVM}        & 75.3            & 80.7          & 84.0           & 87.1             & 81.78            \\ \hline
\textbf{BERT}            & 95.2            & 95.6          & 98.1           & 99.3             & 97.0            \\ \hline
\textbf{Error Reduction} & 80.6\%          & 77.2\%        & 88.1\%         & 94.6\%           & 85.1\%          \\ \hline
\end{tabular}
\bigskip
\caption{Results for the various models on the 6 LexGLUE classification tasks. µF1 refers to micro-averaged F1-score and mF1 to macro-averaged F1 score.\textit{Error reduction is calculated between the Best SVM and the best BERT variant.}}
\hspace*{-\textwidth}\begin{tabular}{|l|ccc|c|cc|cc|cc|c|c|cc|}
\hline
\textbf{Model} & \multicolumn{4}{c|}{\textbf{ECtHR}} & \multicolumn{2}{c|}{\textbf{SCOTUS}} & \multicolumn{2}{c|}{\textbf{EUR-LEX}} & \multicolumn{2}{c|}{\textbf{LEDGAR}} & \multicolumn{2}{c|}{\textbf{Unfair}} & \multicolumn{2}{c|}{\textbf{Average}} \\
 & \multicolumn{2}{c}{\textbf{A}} & \multicolumn{2}{c|}{\textbf{B}} & \multicolumn{1}{l}{} & \multicolumn{1}{l|}{} & \multicolumn{1}{l}{} & \multicolumn{1}{l|}{} & \multicolumn{1}{l}{} & \multicolumn{1}{l|}{} & \multicolumn{2}{c|}{\textbf{Tos}} & \multicolumn{1}{l}{} & \multicolumn{1}{l|}{} \\ \hline
 & \multicolumn{1}{c|}{µF1} & \multicolumn{1}{c|}{mF1} & µF1 & mF1 & \multicolumn{1}{c|}{µF1} & mF1 & \multicolumn{1}{c|}{µF1} & mF1 & \multicolumn{1}{c|}{µF1} & mF1 & µF1 & mF1 & \multicolumn{1}{c|}{µF1} & mF1 \\ \hline
\textbf{SVM} & \multicolumn{1}{c|}{66.3} & \multicolumn{1}{c|}{55.0} & 76.0 & 65.4 & \multicolumn{1}{c|}{74.4} & 64.5 & \multicolumn{1}{c|}{65.7} & 49.0 & \multicolumn{1}{c|}{88} & 82.6 & 95.1 & 75.9 & \multicolumn{1}{l|}{77.6} & \multicolumn{1}{l|}{65.4} \\ \hline
\textbf{BERT} & \multicolumn{1}{c|}{\textbf{71.4}} & \multicolumn{1}{c|}{64} & 79.6 & \textbf{78.3} & \multicolumn{1}{c|}{70.5} & 60.9 & \multicolumn{1}{c|}{71.6} & 55.6 & \multicolumn{1}{c|}{87.7} & 82.2 & 97.3 & 80.4 & \multicolumn{1}{c|}{\textbf{79.7}} & 70.2 \\ \hline
\textbf{\begin{tabular}[c]{@{}l@{}}Legal\\ BERT\end{tabular}} & \multicolumn{1}{c|}{71.2} & \multicolumn{1}{c|}{\textbf{64.9}} & \textbf{80.6} & 77.2 & \multicolumn{1}{c|}{76.2} & 65.8 & \multicolumn{1}{c|}{\textbf{72.2}} & \textbf{56.2} & \multicolumn{1}{c|}{\textbf{88.1}} & \textbf{82.7} & \textbf{97.4} & \textbf{83.4} & \multicolumn{1}{c|}{\textbf{81.0}} & \textbf{71.7} \\ \hline
\textbf{\begin{tabular}[c]{@{}l@{}}CaseLaw\\ BERT\end{tabular}} & \multicolumn{1}{c|}{71.2} & \multicolumn{1}{c|}{64.2} & 79.7 & 76.8 & \multicolumn{1}{c|}{\textbf{76.4}} & \textbf{66.2} & \multicolumn{1}{c|}{71} & 55.9 & \multicolumn{1}{c|}{88} & 82.3 & \textbf{97.4} & 82.4 & \multicolumn{1}{c|}{80.6} & 71.3 \\ \hline
\textbf{\begin{tabular}[c]{@{}l@{}}Error\\ Red. \%\end{tabular}} & \multicolumn{1}{c|}{15.1} & \multicolumn{1}{c|}{22} & 19.2 & 37.3 & \multicolumn{1}{c|}{7.8} & 4.8 & \multicolumn{1}{c|}{19.0} & 14.1 & \multicolumn{1}{c|}{0.8} & 0.6 & 46.9 & 31.1 & \multicolumn{1}{c|}{18.1} & 18.3 \\ \hline
\end{tabular}\hspace*{-\textwidth}
\end{table}

\section{Discussion}

The results highlight an interesting phenomenon: while the average micro-F1 error reduction on general domain benchmarks is \textbf{85.1\%}, it is only \textbf{18.1\%} on legal text classification tasks.\\
It is worth noting that the performance increases for BERT-based models is larger on the LexGLUE tasks with more pronounced label imbalance, with just a few labels representing the vast majority of a dataset and certain classes having very few examples. This is also noticeable with a much larger increase in macro-F1 score when compared to the increase in micro-F1 score on the same tasks. This highlights the increased robustness of the BERT-based models, whose use of pre-training and transfer learning techniques allow them to reach certain levels of performance performance with much fewer examples than traditional methods such as SVM. As training in the legal domain can be difficult to acquire and label, this is an important step towards democratising legal NLP.\\

Despite this and their impressive performance in both the general domain and other specialised domains, it is still apparent that in the legal domain, BERT-based models, even with specific domain pre-training, produce only a modest improvement on the 6 evaluated LexGLUE tasks.

There is no clear explanation for this phenomenon, but we discuss multiple hypotheses (a, b, c).
The first (\textbf{a}), initially proposed by Zheng et al. \cite{legalberticail} to explain the mild improvements from Legal-BERT, is that the tasks on which we evaluate legal NLP algorithms are not suitable, either due to them being too simple or their language not being sufficiently domain-specific to take advantage of the models' pretraining. However, this does not provide a clear explanation for the overall weak improvement from deep learning over SVM classifiers. 

A similar potential explanation (\textbf{b}) could be that \textit{simple} mono-lingual text classification is not enough to truly take advantage of the possibilities offered by more powerful BERT-based models. This would indicate that the powerful language representation of Legal-BERT models could be key to tackling more complex tasks. Such tasks have now started being explored, such as legal rationale extraction \cite{rationale} or textual entailment in the form of a multiple choice QA task \cite{legalberticail}. This QA task, \textit{CaseHold}, is the final task of the LexGLUE benchmark, and shows a noticeably better increase in performance of Legal BERT models over General Domain BERT models, which supports the interest of further studies towards validating this hypothesis.

However, this explanation does not fully address the weak performance gains on text classification. A final hypothesis (\textbf{c}) we propose is that large language models, even when trained on legal language, still lack the ability to capture the depth of legal language and its specific vocabulary. These models could also fail to properly weigh the meaning of multiple legal concepts being mentioned together. This hypothesis would suggest the need to develop a way to integrate sources of legal information, such as  knowledge-bases or ontologies, within deep learning models to truly take advantage of their potential.\\

\section{Conclusion and Future Work}
We experimentally confirm the intuition that SVM classifiers perform remarkably well on multiple legal text classification benchmarks. We notably highlight that the relative performance improvement between BERT-based models and SVM models is considerably smaller within the legal domain than on general domain classification tasks, even with BERT models specifically trained for the legal domain.


We propose and discuss three potential explanations for these results. Future work will focus on exploring the limits of BERT models within the legal field, both by exploring more complex tasks and integrating existing knowledge bases with them.

We believe our results do not indicate the unsuitability of BERT-based approaches, but rather show that they have shortcomings and that they are perhaps better suited to more complex tasks. We also show that in the case of imbalanced datasets with very few examples for some of their classes, BERT-based models result in bigger increases in macro F1-score than in micro-F1 score, showcasing their ability to reach better results in low-data downstream tasks.

Future work will focus on exploring and pushing the limits of BERT models (and variants) within the legal domain, both by exploring more complex tasks and attempting to integrate external knowledge bases within them to improve performance on tasks such as text classification.

We hope our work will help support future work on Legal NLP focusing on exploring the specificities of legal text and better taking them into account. We make our experiments' code available to support future work.

\section{Acknowledgements}
This work was granted access to the HPC resources of IDRIS under the allocation AD011012667 made by GENCI.

Many thanks to Paul Briton, Rym Laabiyad, Akshita Gheewala and Francesco Piccoli for their advice on this paper.

\bibliographystyle{vancouver}
\bibliography{nbsvm}

\end{document}